\documentclass[runningheads,a4paper]{llncs}

\usepackage{amssymb}
\usepackage{graphicx}
\usepackage{algorithm,algorithmic}
\usepackage{amsmath,amssymb,amscd,amstext}
\usepackage{tikz}
\usetikzlibrary{arrows}
\usepackage{subcaption}
\usepackage{multirow,url}
\usetikzlibrary{intersections,positioning, fit, calc, shapes, arrows}

\usepackage{url}
\urldef{\mailsa}\path|lgondara@sfu.ca,wangk@cs.sfu.ca|

\begin{document}

\mainmatter  

\title{MIDA: Multiple Imputation using Denoising Autoencoders}

\titlerunning{MIDA: Multiple Imputation using Denoising Autoencoders}

%
%
\author{Lovedeep Gondara\and Ke Wang}
\authorrunning{Lovedeep Gondara\and Ke Wang}

\institute{Department of Computing Science\\
Simon Fraser University\\
\mailsa\\}

%
%

\toctitle{MIDA: Multiple Imputation using Denoising Autoencoders}
\tocauthor{Gondara and Wang}
\maketitle

\begin{abstract}
Missing data is a significant problem impacting all domains. State-of-the-art framework for minimizing missing data bias is multiple imputation, for which the choice of an imputation model remains nontrivial. We propose a multiple imputation model based on overcomplete deep denoising autoencoders. Our proposed model is capable of handling different data types, missingness patterns, missingness proportions and distributions. Evaluation on several real life datasets show our proposed model significantly outperforms current state-of-the-art methods under varying conditions while simultaneously improving end of the line analytics.
\end{abstract}

\section{Introduction}
Missing data is an important issue, even small proportions of missing data can adversely impact the quality of learning process, leading to biased inference \cite{rubin1976inference,chen2008optimization}. Many methods have been proposed over the past decades to minimize missing data bias \cite{rubin1976inference,little1988missing} and can be divided into two categories: One that attempt to model the missing data process and use all available partial data for directly estimating model parameters and two that attempt to fill in/impute missing values with plausible predicted values. Imputation methods are preferred for their obvious advantage, that is, providing users with a complete dataset that can be analyzed using user specified models.

Methods for imputing missing data range from replacing missing values by the column average to complex imputations based on various statistical and machine learning models. All standalone methods share a common drawback, imputing a single value for one missing observation, which is then treated as the gold standard, same as the observed data in any subsequent analysis. This implicitly assumes that imputation model is perfect and fails to account for error/uncertainty in the imputation process. This is overcome by replacing each missing value with several slightly different imputed values, reflecting our uncertainty about the imputation process. This approach is called \emph{multiple imputation} \cite{little2014statistical,schafer1999multiple} and is the most widely used framework for missing data analytics. The biggest challenge with multiple imputation is the correct specification of an imputation model \cite{morris2014tuning}. It is a nontrivial task because of the varying model capabilities and underlying assumptions. Some imputation models are incapable of handling mixed data types (categorical and continuous), some have strict distributional assumptions (multivariate normality) and/or cannot handle arbitrary missing data patterns. Existing models capable of overcoming aforementioned issues are further limited in their ability to model highly nonlinear relationships, high volume data and complex interactions while preserving inter-variable dependencies.

Recent advancements in deep learning have established state-of-the-art results in many fields \cite{lecun2015deep}. Deep architectures have the capability to automatically learn latent representations and complex inter-variable associations, which is not possible using classical models. Part of the deep learning framework, Denoising Autoencoders (DAEs) \cite{vincent2008extracting} are designed to recover clean output from noisy input. Missing data is a special case of noisy input, making DAEs ideal as an imputation model. But, missing data  can depend on interactions/latent representations that are not observable in the input dataset space. Hence, we propose to use an \emph{overcomplete} DAE as an imputation model, whereby projecting our input data to a higher dimensional subspace from where we then recover missing information. We propose a multiple imputation framework with overcomplete DAE as the base model, where we simulate multiple predictions by initializing our model with a different set of random weights at each run. Details of our method are presented in Section \ref{sec:our_model}. Our proposed method has several advantages over the current methods, some of which we outline below:
\begin{itemize}
    \item Previous studies on imputing missing data using machine learning methods use complete observations for the training phase. We show that our model outperforms state-of-the-art methods even when users do not have the luxury of having complete observations for initial training, a common scenario in real life.
    \item Our model is capable of preserving attribute correlations, which are of a concern using traditional imputation methods and can significantly affect end of the line analytics
    \item Our model is better equipped to deal with different missing data generation processes, such as data missing not at random, which is a performance bottleneck for other imputation methods. Experimental results using real life datasets show that our model outperforms state-of-the-art methods under varying dataset and missingness conditions and improves end of the line analytics. 
\end{itemize}
The rest of the paper is organized as following. Section \ref{sec:prelim} provides  preliminary background to missing data terminology and introduces denoising autoencoders. Section \ref{sec:our_model} introduces our model with section \ref{sec:eval} presenting empirical evaluation and the effect of imputation on end of the line analytics followed by our conclusions.

\section{Background}\label{sec:prelim}
Missing data is a well researched topic in statistics. Most of the early work on missing data, including definitions, multiple imputation and subsequent analysis is attributed to works of Little and Rubin \cite{little2014statistical,little1988missing,rubin1976inference}. From machine learning perspective, it has been shown that  auto-associative neural networks are better at imputing missing data when attribute interdependencies are of concern \cite{nelwamondo2007missing}, a common scenario in real life datasets. Denoising autoencoders have been recently used in completing traffic and health records data \cite{duan2014deep,beaulieu2016missing} and collaborative filtering \cite{li2015deep}. Below we provide some preliminary introduction to missing data mechanisms and denoising autoencoders.

\subsection{Missing data} 
\textbf{Mechanisms:} Impact of missing data depends on the underlying missing data generating mechanism. We define three missing data categories \cite{little2014statistical} with the aid of data from Table \ref{dummy_data}, representing an income questionnaire in a survey where we denote missing data with "?".
\begin{table}[]
\centering
\caption{Data snippet for income questionnaire with missing data represented using '?'}
\label{dummy_data}
\begin{tabular}{|l|l|l|l|l|l|l|}
\hline
Id & Age & Sex & Income & Postal & Job & Marital status \\ \hline
1  & 50  & M   & 100    & 123    & a   & single         \\ \hline
2  & 45  & ?   & ?      & 456    & ?   & married         \\ \hline
3  & ?   & F   & ?      & 789    & ?   & ?              \\ \hline
\end{tabular}
\end{table}
Data is Missing Completely At Random (\emph{MCAR}) if missingness does not depend on observed or unobserved data, example: Survey participants flip a coin to decide whether to answer questions or not. Data is Missing At Random (\emph{MAR}) if missingness can be explained using observed data, example: Survey participants that live in postal code 456 and 789 refuse to fill in the questionnaire. Data is Missing Not At Random (\emph{MNAR}) if missingness depends on an unobserved attribute or on the missing attribute itself, example: Everyone who owns a six bedroom house refuses the questionnaire, bigger house is an indirect indicator for greater wealth and a better paying job, but we don't have the related data. When data are MAR or MCAR, it is known as ignorable missing data as observed data can be used to account for missingness. But, given the observed data, it is impossible to distinguish between MNAR and MAR \cite{sterne2009multiple} and sometimes, missing data can be a combination of both.\\
\textbf{Multiple Imputation:} In a multiple imputation scenario, we will create multiple copies of the dataset presented in Table \ref{dummy_data} with '?' replaced by slightly different imputed values in each copy. Multiple imputation accounts for uncertainty in predicting missing data using observed data by modelling variability into the imputed values as the true values for missing data are never known. Multiple imputed datasets are then analyzed independently and the results combined. A single statistic such as classification accuracy or root mean square error (RMSE) can be simply averaged from multiple imputations.

\subsection{Autoencoders and Denoising autoencoders}
An autoencoder takes an input $x \in [0,1]^d$ and maps (encodes) it to an intermediate representation $h \in [0,1]^{d'}$ using an encoder, where $d'$ represents a different dimensional subspace. The assumption is, in the dataset, $h$ captures the coordinates along the main factors of variation. The encoded representation is then decoded back to the original $d$ dimensional space using a decoder. Encoder and decoder are both artificial neural networks. The two stages are represented as
\begin{equation}
    h=s(Wx+b)
\end{equation}
\begin{equation}
    z=s(W'h+b')
\end{equation}
where $z$ is the decoded result and $s$ is any nonlinear function, reconstruction error between $x$ and $z$ is minimized during training phase. 

Denoising autoencoders, are a natural extension to autoencoders \cite{vincent2008extracting}. By corrupting the input data and forcing the network to reconstruct the clean output forces the hidden layers to learn robust features. Corruption can be applied in different ways, such as randomly setting some input values to zero or using distributional additive noise. DAEs reconstruction capabilities can be explained by thinking of DAEs implicitly  estimating  the  data  distribution  as  the  asymptotic  distribution  of  the  Markov  chain  that alternates between corruption and denoising \cite{bengio2013generalized}.

\section{Models} \label{sec:our_model}
This section introduces our multiple imputation model and the competitors used for comparison.

\subsection{Our model}
\begin{figure}[t]
\centering
\begin{tikzpicture}[
    circ/.style={circle, minimum size=0.9cm, 
        draw=black!70, fill=white},
    cross/.style={circle, minimum size=0.9cm,
        draw=black!70, fill=white,
        path picture={\node[cross out, draw, ultra thick, red, minimum size=5mm]{};}},
    x1/.style={circle, minimum size=0.9cm,
        draw=black!70, fill=white,
        path picture={\node[]{$x_1$};}},
    x1p/.style={circle, minimum size=0.9cm,
        draw=black!70, fill=white,
        path picture={\node[]{$x_1'$};}},
    xn/.style={circle, minimum size=0.9cm,
        draw=black!70, fill=white,
        path picture={\node[]{$x_n$};}},
    xnp/.style={circle, minimum size=0.9cm,
        draw=black!70, fill=white,
        path picture={\node[]{$x_n'$};}},
    xn7/.style={circle, minimum size=0.9cm,
        draw=black!70, fill=white,
        path picture={\node[]{$x_{n+\Theta}'$};}},
    xn14/.style={circle, minimum size=0.9cm,
        draw=black!70, fill=white,
        path picture={\node[]{$x_{n+2\Theta}'$};}},
    xn21/.style={circle, minimum size=0.9cm,
        draw=black!70, fill=white,
        path picture={\node[]{$x_{n+3\Theta}'$};}},
    H/.style={rounded corners, fill=black!20, row sep=2mm},
    column/.pic={
        \matrix (#1) [H, label={[name=l#1]#1}]
        {\node[x1]{};\\
        \node[cross]{};\\
        \node[circ]{};\\
        \node[cross]{};\\
        \node[cross]{};\\
        \node[xn]{};\\};
        },    
    column2/.pic={
        \matrix (#1) [H, label={[name=l#1]#1}]
        {\node[x1p]{};\\
        \node[circ]{};\\
        \node[circ]{};\\
        \node[circ]{};\\
        \node[circ]{};\\
        \node[xn7]{};\\};
        },
    column3/.pic={
        \matrix (#1) [H, label={[name=l#1]#1}]
        {\node[x1p]{};\\
        \node[circ]{};\\
        \node[circ]{};\\
        \node[circ]{};\\
        \node[circ]{};\\
        \node[xn14]{};\\};
        },
    column4/.pic={
        \matrix (#1) [H, label={[name=l#1]#1}]
        {\node[x1p]{};\\
        \node[circ]{};\\
        \node[circ]{};\\
        \node[circ]{};\\
        \node[circ]{};\\
        \node[xn21]{};\\};
        },
   column5/.pic={
        \matrix (#1) [H, label={[name=l#1]#1}]
        {\node[x1p]{};\\
        \node[circ]{};\\
        \node[circ]{};\\
        \node[circ]{};\\
        \node[circ]{};\\
        \node[xnp]{};\\};
        }
    ]

    \path (0,0) pic{column=I};
    \path (1.5,0) pic{column2=H1};
    \path (3,0) pic{column3=H2};
    \path (4.5,0) pic{column4=H3};

    \node[fit=(I) (lI) (H2) (H3), draw, label=Encoder]{};

    \begin{scope}[xshift=6.5cm]
    \path (0,0) pic{column4=H3};
    \path (1.5,0) pic{column3=H4};
    \path (3,0) pic{column2=H5};
    \path (4.5,0) pic{column5=O};

    \node[fit=(H3) (H4) (lO) (O), draw, label=Decoder]{};
    \end{scope}

\end{tikzpicture}
	\caption{Our basic architecture, encoder block increases dimensionality at every hidden layer by adding $\Theta$ units with decoder symmetrically scaling it back to original dimensions. Crossed out inputs represent stochastic corruption at input by setting random inputs to zero. $H_1, H_2, H_3, H_4, H_5$ are hidden layers with $I$ and $O$ being the input and output layers respectively. Encoder and decoder are constructed using fully connected artificial neural networks.}\label{daearch}
\end{figure}
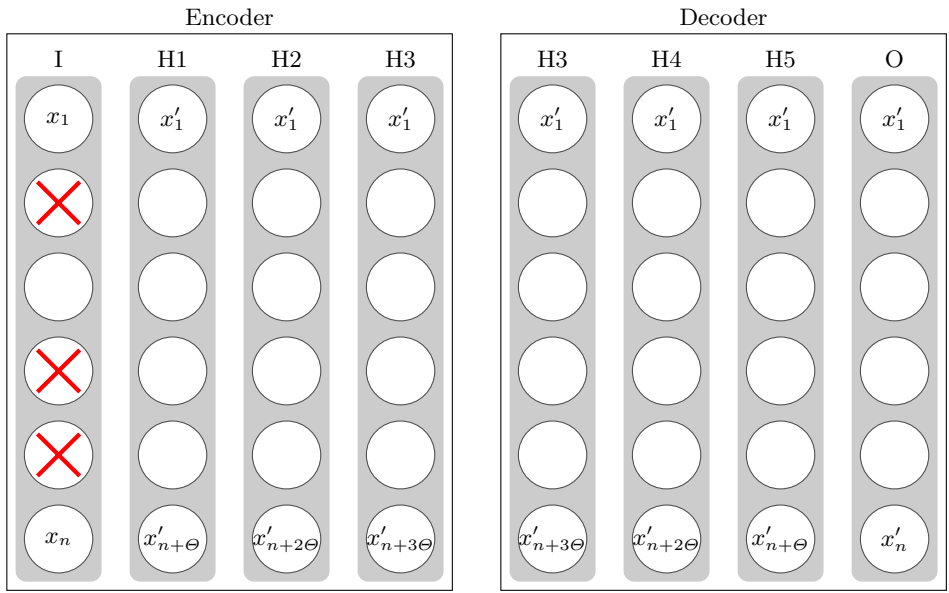
\textbf{Architecture:}Our default architecture is shown in Figure \ref{daearch}. We employ atypical overcomplete representation of DAEs, that is, more units in successive hidden layers during encoding phase compared to the input layer. This mapping of our input data to a higher dimensional subspace creates representations capable of adding lateral connections, aiding in data recovery, usefulness of this approach is empirically validated in the supplemental material.  We start with an initial  $n$ dimensional input, then at each successive hidden layer, we add $\Theta$ nodes, increasing the dimensionality to $n+\Theta$. For initial comparisons, we use $\Theta=7$. We tried different values for $\Theta$ for various datasets and decided to use 7 as it provided consistent better results. It is an arbitrary choice and can be dealt with by viewing $\Theta$ as another tuning hyperparameter. Our model inputs are standardized between 0 and 1 to facilitate faster convergence for small to moderate sample sizes. Our model is trained with 500 epochs using an adaptive learning rate with a time decay factor of 0.99 and Nesterov's accelerated gradient \cite{nesterov1983method}. The input dropout ratio to induce corruption is set to 0.5, so that in a given training batch, half of the inputs are set to zero. Tanh is used as an activation function as we find it performs better than ReLU for small to moderate sample sizes when some inputs are closer to zero. We use early stopping rule to terminate training if desired mean squared error (MSE) of 1e-06 is achieved or if simple moving average of length 5 of the error deviance does not improve. The training-test split of 70-30 is used with all results reported on the test set. Multiple imputation is accomplished by using multiple runs of the model with a different set of random initial weights at each run. This provides us with the variation needed for multiple imputations. Algorithm \ref{mi_algo} explains our multiple imputation process.\\
\begin{algorithm}
\caption{Multiple imputation using DAEs}
\label{mi_algo}
\begin{algorithmic}[1]
{
\REQUIRE
         $k$: Number of imputations needed
\FOR{$i=1 \to k$}
  
    \STATE Initialize DAE based imputation model using weights from random uniform distribution
    \STATE Fit the imputation model to training partition using stochastic corruption 
    \STATE Reconstruct test set using the trained model
\ENDFOR
}
\end{algorithmic}
\end{algorithm}
\textbf{Usage:}We start with imitating a real life scenario, where the user only have a dataset with pre-existing missing values. That is, the user does not have the luxury of access to the clean data and user does not know the underlying missing data generating mechanism or the distribution. In scenarios where missingness is inherent to the datasets, training imputation models using complete data can bias the learner. But as DAEs require complete data at initialization, we initially use the respective column average in case of continuous variables and most frequent label in case of categorical variables as placeholders for missing data at initialization. Training phase is then initiated with a stochastic corruption process setting random inputs to zero, where our model learns to map corrupted input to clean output. Our approach is based on one assumption, that is, we have enough complete data to train our model, so the model learns to recover \emph{true} data using stochastic corruption on inputs, and is not learning to map placeholders as valid imputations. The results show this assumption is readily satisfied in real life scenarios, even datasets with small sample sizes are enough for DAE based imputation model to achieve better performance compared to state-of-the-art. 

\subsection{Competitors and comparison}
\textbf{Competitor:}For multiple imputation, we need methods that can inject variation in successive imputations, providing slightly different imputation results at each iteration. Simple models such as linear/logistic regression or deterministic methods based on matrix decomposition fail to take this into account. Current state-of-the-art in multiple imputation is the Multivariate Imputation by Chained Equations (MICE) \cite{buuren2011mice}, which is a fully conditional specification approach and works better than Joint Modelling approaches where multivariate distributions cannot provide a reasonable description of the data or where no suitable multivariate distributions can be found. MICE specifies multivariate model on variable by variable basis using a set of conditional densities, one for each variable with missing data. MICE draws imputations by iterating over conditional densities, it has an added advantage of being able to model different densities for different variables. Internal imputation model in MICE is vital and a model with properties to handle different data types and distributions is essential for effective imputations. Predictive mean matching and Random Forest are the best available options within MICE framework \cite{shah2014comparison}. We compared them both and found predictive mean matching to provide more consistent results with varying dataset types and sizes. Hence it is used as the internal component of our competitor MICE model.\\
\textbf{Comparison}: Imputation results are compared using sum of root mean squared error calculated per attribute on the test set, given as
\begin{equation}
    RMSE_{sum}=\sum_{i=1}^m \sqrt{E(\sum_{i=1}^n (\hat{t_i} - t_i))}
\end{equation}
where we have $m$ attributes, $n$ observations, $\hat{t}$ is the imputed value and $t$ is the observed value. $RMSE_{sum}$ is calculated on scaled datasets to avoid disproportionate attribute contributions. $RMSE_{sum}$ provides us a measure of relative distance, that is, how far the dataset completed with imputed values is from the original complete dataset. For multiple imputation scenarios with $k$ imputations, we have $k$ values for $RMSE_{sum}$ per dataset. The results are then reported using average $RMSE_{sum}$ along with the range.

\section{Experiments} \label{sec:eval}
We start our empirical evaluation for multiple imputation on several publicly available real life datasets under varying missingness conditions.

\subsection{Datasets}
Table \ref{datasets} shows the properties of various real life publicly available datasets \cite{leisch2010machine} used for model evaluation. Models based on deep architectures are known to perform well on large sample, high dimensional datasets. Here we include some extremely low dimensional and low sample size datasets to test the extremes and to prove that our model has real world applications. Most of the datasets have a combination of continuous, categorical and  ordinal attributes, which further challenges the convergence of our model using small training samples.

\begin{table}[]
\centering 
\caption{Datasets used for evaluation. Dataset acronyms are shown in parenthesis that we will be using in the results section.}
\label{datasets}
\begin{tabular}{|l|l|l|}
\hline
               & Observations & Attributes \\ \hline
Boston Housing (BH) & 506        & 14           \\ \hline
Breast Cancer (BC)  & 699        & 11           \\ \hline
DNA (DN)            & 3186       & 180          \\ \hline
Glass (GL)         & 214        & 10           \\ \hline
House votes (HV)   & 435        & 17           \\ \hline
Ionosphere (IS)    & 351        & 35           \\ \hline
Ozone (ON)         & 366        & 13           \\ \hline
Satellite (SL)     & 6435       & 37           \\ \hline
Servo (SR)         & 167        & 5            \\ \hline
Shuttle (ST)       & 58000      & 9            \\ \hline
Sonar (SN)         & 208        & 61           \\ \hline
Soybean (SB)       & 683        & 36           \\ \hline
Vehicle (VC)       & 846        & 19           \\ \hline
Vowel (VW)         & 990        & 10           \\ \hline
Zoo (ZO)           & 101        & 17           \\ \hline
\end{tabular}
\end{table}

\subsection{Inducing missingness}\label{introducemissing}
To provide a wide range of comparisons, initially for each data set, we introduce missingness in four different ways, with a fixed missingness proportion of 20\% using the steps detailed below.
\begin{enumerate}
    \item Append a uniform random vector $v$ with $n$ observations to the dataset with values between 0 and 1, where $n$ is number of observations in the dataset.
    \item \emph{MCAR, uniform}: Set all attributes to have missing values where $v_i \le t$, , $i \in 1:n$, $t$ is the missingness threshold, 20\% in our case.
    \item \emph{MCAR, random}: Set randomly sampled half of the attributes to have missing values where $v_i \le t$, $i \in 1:n$.
    \item \emph{MNAR, uniform}: Randomly sample two attributes $x_1$ and $x_2$ from the dataset and calculate their median $m_1$ and $m_2$. Set all attributes to have missing values where $v_i \le t$, $i \in 1:n$ and ($x_1 \le m_1$ or $x_2 \ge m2$).
    \item \emph{MNAR, random}: Randomly sample two attributes $x_1$ and $x_2$ from the dataset and calculate their median $m_1$ and $m_2$. Set randomly sampled half of the attributes to have missing values where $v_i \le t$, $i \in 1:n$ and ($x_1 \le m_1$ or $x_2 \ge m2$).
\end{enumerate}

\subsection{Main results}
Table \ref{mainresult} shows the multiple imputation results on real life datasets, comparing five imputations by our model with five imputations by MICE, that is, each missing value is imputed five times with a slightly different value.
\begin{table}[!h]
\centering
\caption{Imputation results comparing our model and MICE. Results are displayed using sum of root mean square error($RMSE_{sum}$), providing a measure of relative distance of imputation from original data. As results are from multiple imputation (5 imputations), mean $RMSE_{sum}$ from 5 imputations is displayed outside with min and max $RMSE_{sum}$ inside parenthesis providing a range for imputation performance. Value for MNAR is NA for dataset VW as MICE was unable to impute a complete dataset.}
\label{mainresult}
\begin{tabular}{|l|l|l|l|l|l|}
\hline
                       & Data & \multicolumn{2}{l|}{\phantom{epmtyspacee}Uniform missingness}            & \multicolumn{2}{l|}{\phantom{epmtyspaee}Random missingness}            \\ \hline
                       &      & DAE                & MICE                & DAE                & MICE               \\ \hline
\multirow{15}{*}{MCAR} & BH      & 2.9(2.9,3)                       & 3.7(3.5,3.8)                      & 0.9(0.9,1)                              & 0.9(0.7,1) \\ \cline{2-6}
                       &  BC    & 2.9(2.9,2.9)                      & 3.9(3.6,4.2)                      & 1.2(1.2,1.3)                   & 1.3(1.1,1.4)                   \\ \cline{2-6}
                       & DN      & 25.7(25.7,25.7)                  & 36.5(36.3,36.6)                   & 13.1(13.1,13.2)                & 16.9(16.9,17)                  \\ \cline{2-6}
                       & GL     & 1.1(1,1.1)                        & 1.5(1.3,1.7)                      & 1.3(1.2,1.4)                   & 1.4(1.3,1.6)          \\ \cline{2-6}
                       & HV      & 2.4(2.4,2.4)                     & 3.4(3.1,3.7)                      & 1.1(1.1,1.2)                  & 1.2(0.9,1.3)                    \\ \cline{2-6}
                       & IS     & 13(12.9,13.1)                     & 17.1(16.2,17.7)                   & 5.8(5.6,6.2)                   &  7(6.7,7.5)                 \\ \cline{2-6}
                       & ON     & 2.1(2.1,2.1)                      & 3.1(3,3.3)                        & 0.9(0.9,1)                     & 1(1,1.2)                  \\ \cline{2-6}
                       & SL     & 3.6(3.6,3.7)                      & 4.5(4.4,4.6)                      & 1.8(1.7,1.8)                            & 0.7(0.7,0.7)                  \\ \cline{2-6}
                       & SR     & 1.2(1.,1.2)                      & 1.5(1.4,1.7)                      & 0.4(0.4,0.5)                            & 0.4(0.4,0.5)                 \\ \cline{2-6}
                       & ST     & 16.5(16.5,16.7)                   & 27.9(27.5,28.2)                   & 6.5(6.4,6.7)                   & 13(12.5,13.8)              \\ \cline{2-6}
                       & SN     & 5.1(5,5.1)                        & 7.3(7.2,7.5)                      & 2.3(2.2,2.3)                   & 3.2(3.2,3.3)                   \\ \cline{2-6}
                       & SB     & 1.8(1.8,1.8)                      & 2.4(2.3,2.4)                      & 1.2(1.1,1.2)                            & 1.1(1,1.1)                  \\ \cline{2-6}
                       & VC     & 4.1(4,4.1)                        & 5.6(5.5,5.7)                      & 1.6(1.6,1.6)                   & 2.2(2.1,2.3)                  \\ \cline{2-6}
                       & VW     & 5.8(5.7,6.2)                      & 7.7(7,8.1)                        & 2.6(2.4,2.9)                   & 3.8(3.3,4.2)                  \\ \cline{2-6}
                       & ZO     & 2.1(2.1,2.1)                      & 3.4(3.1,4.3)                      & 1.1(1.1,1.2)                            & 1.1(1.1,1.1)                   \\ \hline
\multirow{15}{*}{MNAR} & BH     & 2.3(2.2,2.4)                      & 3.2(2.9,3.4)                      & 0.9(0.8,1)                     & 0.7(0.7,0.8)                  \\ \cline{2-6}
                       & BC      & 2.9(2.8,3)                       & 3.6(3.4,3.8)                      & 1.7(1.7,1.8)                            & 1.4(1.3,1.5)                  \\ \cline{2-6}
                       & DN     & 25.3(25.2,25.3)                   & 34.5(34.5,34.7)                   & 5.7(5.7,5.8)                   & 7.2(7.1,7.2)                   \\ \cline{2-6}
                       & GL     & 1.3(1.3,1.4)                      & 1.5(1.3,1.8)                      & 0.4(0.3,0.4)                            & 0.2(0.11,0.2)                  \\ \cline{2-6}
                       & HV      & 2.6(2.6,2.6)                     & 3.5(3.3,3.7)                      & 1.3(1.2,1.3)                            & 1.3(1.3,1.4)                  \\ \cline{2-6}
                       & IS     & 11.7(11.5,11.8)                   & 15.4(14.9,16.5)                   & 4.8(4.5,5.1)                   & 6.3(5.6,6.8)                  \\ \cline{2-6}
                       & ON     & 1.5(1.5,1.5)                      & 2.2(2,2.4)                        & 1.2(1.1,1.2)                   & 1.3(1.1,1.5)                   \\ \cline{2-6}
                       & SL     & 3.4(3.4,3.4)                      & 3.8(3.8,3.9)                      & 1.6(1.6,1.6)                            & 0.5(0.5,0.5)                   \\ \cline{2-6}
                       & SR     & 1.2(1.2,1.2)                      & 1.6(1.5,1.7)                      & 0.4(0.3,0.4)                            & 0.3(0.2,0.3)                  \\ \cline{2-6}
                       & ST     & 11.8(11.7,11.9)                   & 22.4(22.1,22.7)                   & 4.5(4.3,4.7)                   & 9.5(8.4,10.3)                  \\ \cline{2-6}
                       & SN     & 4.6(4.6,4.6)                      & 6.8(6.5,7.1)                      & 2.3(2.3,2.4)                   & 3.1(3,3.2)                  \\ \cline{2-6}
                       & SB     & 1.7(1.7,1.7)                      & 2.3(2.2,2.4)                      & 0.6(0.6,0.6)                   & 0.9(0.9,0.9)                  \\ \cline{2-6}
                       & VC     & 3.5(3.4,3.7)                      & 4.6(4.4,4.8)                      & 1.7(1.7,1.8)                   & 2.4(2.3,2.4)                  \\ \cline{2-6}
                       & VW     & 5.9(5.9,5.9)                      & NA                                & 2.3(2.1,2.5)                    & NA                 \\ \cline{2-6}
                       & ZO     & 3.3(2.8,5.5)                      & 3.9(3.6,4.6)                      & 0.9(0.8,1.0)                   & 1.1(0.7,1.7)                   \\ \hline
\end{tabular}
\end{table}
The results show that our model outperforms MICE in $100\%$ of cases with data MCAR and MNAR with uniform missing pattern and in $>70\%$ of cases with random missing pattern. Our model's superior performance in this scenario using small to moderate dataset sizes with constrained dimensionality is indicative of it's utility when datasets are large and are of higher dimensionality, which is a performance bottleneck for other multiple imputation models whereas our model is capable of handling massive data by design. Another advantage is that our model does not need a certain proportion of available data to predict missing value. As in the case of dataset VW-MNAR, MICE was unable to provide complete imputations.

Computational cost associated with our model is at par or better than imputations based on MICE for small to moderate sized datasets. This might seem counter-intuitive to some readers as our model is much more complex. But, computational gains are significant when we are modelling a complete dataset in a single attempt compared to iterative variable by variable imputation in MICE.

\subsection{Increased missingness proportion}
Missing data proportion is known to affect imputation performance, which deteriorates with increasing missing data proportion. To test the impact of varying missing data proportion on our model, we introduce missingness in all 15 datasets with missingness proportion set at 40\% and 60\% using methods described in experimental setup section. Keeping all model parameters same for our model and MICE, we multiple imputed datasets with five imputations each. For a better visual representation, we compare the imputation results between our model and MICE using mean error ratio $E_R$, given as
\begin{equation}
    E_R = \dfrac{\dfrac{1}{n}\sum_{i=1}^n E_{Di}}{\dfrac{1}{n}\sum_{i=1}^n E_{Pi}}
\end{equation}
where $E_D$ is imputation error of our model, $E_P$ is imputation error of MICE and $n$ is number of imputations. $E_R$ values of less than one signify average superior performance of our model over MICE, whereas values greater than one signify MICE performing better. 

\begin{figure}[t]
\centering
\begin{tabular}{cccc}
\subcaptionbox{}{\includegraphics[scale=0.15]{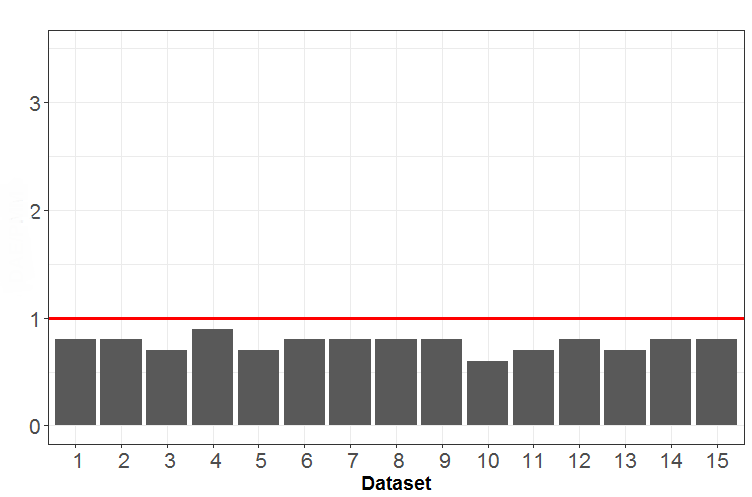}} &
\subcaptionbox{}{\includegraphics[scale=0.15]{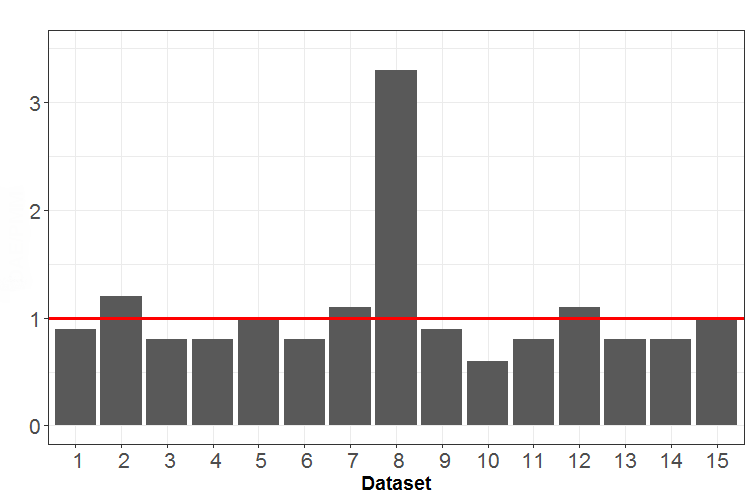}} &
\subcaptionbox{}{\includegraphics[scale=0.15]{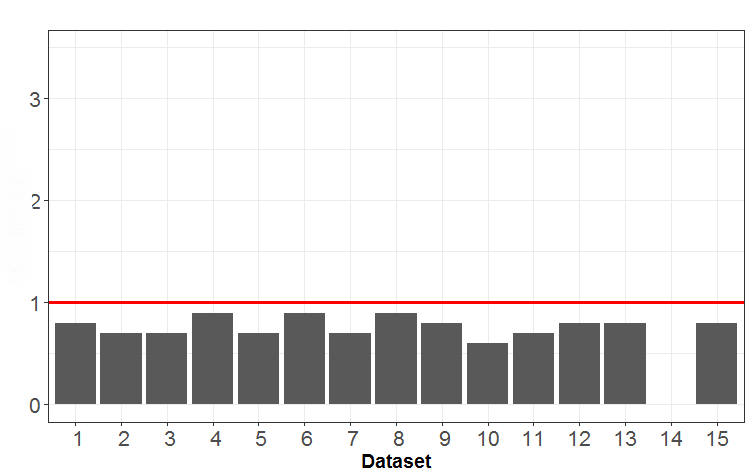}} &
\subcaptionbox{}{\includegraphics[scale=0.15]{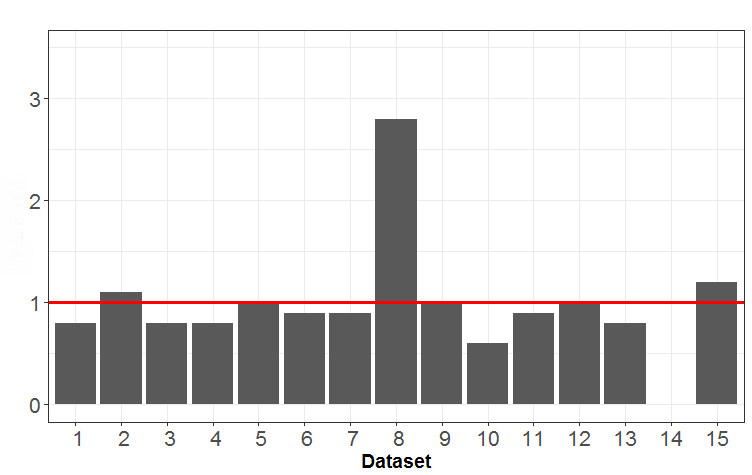}} \\
\subcaptionbox{}{\includegraphics[scale=0.15]{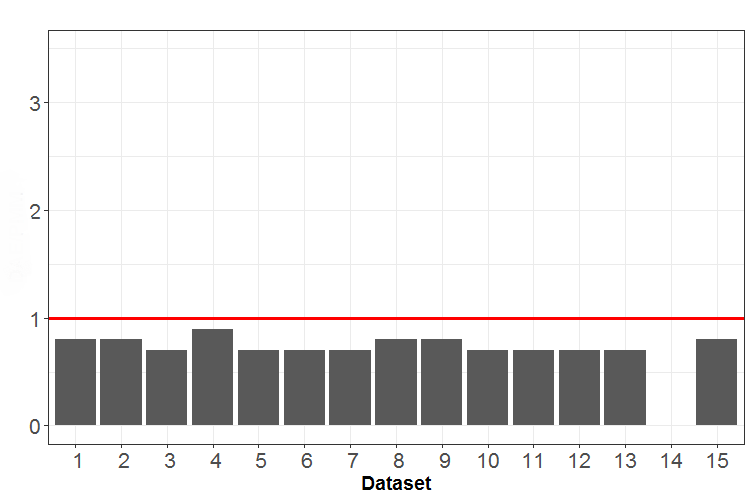}} &
\subcaptionbox{}{\includegraphics[scale=0.15]{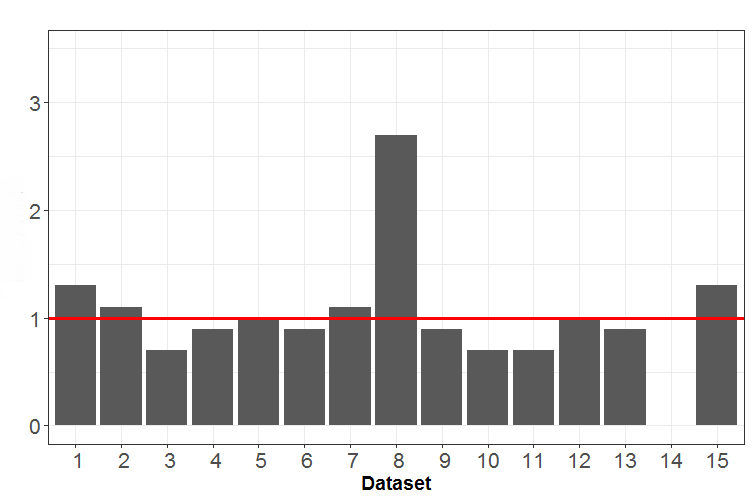}} &
\subcaptionbox{}{\includegraphics[scale=0.15]{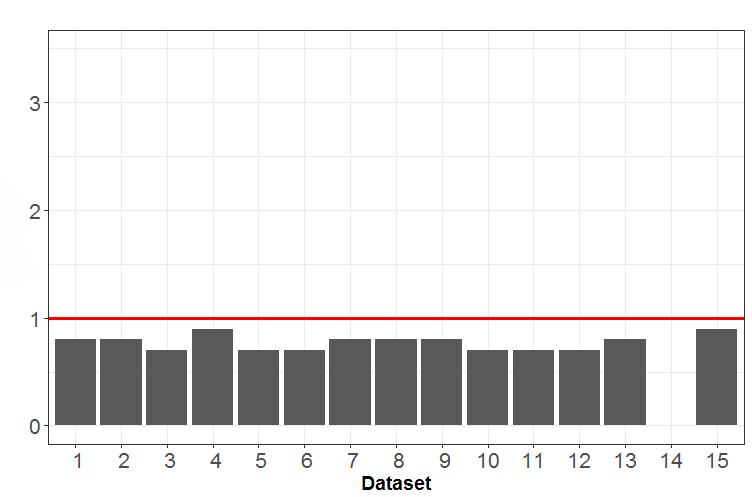}} &
\subcaptionbox{}{\includegraphics[scale=0.15]{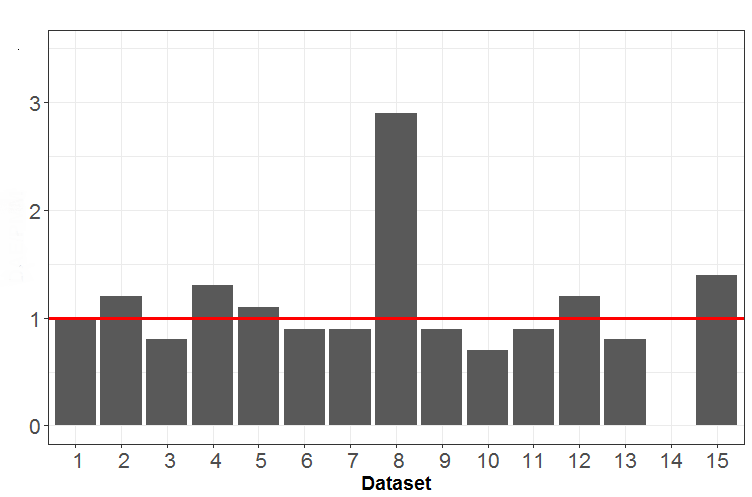}} 
\end{tabular}
\caption{Results for imputation with increased missingness proportions. Figures (a), (b), (c), (d) show imputation results with 40\% missing data and figures (e), (f), (g), (h) show imputation results with 60\% missing data. Red line is drawn as a reference line at y-intercept of 1 to signify superior/inferior performance of our model vs MICE. Results are displayed using $E_R$ where values less than one signify our model performing better and values greater than one signify MICE's superior performance. X-axis show different datasets (1-15) and Y-axis display $E_R$. For some cases, MICE has trouble imputing dataset 14 (VW) whereas our model provides consistent imputations. }
\label{increased_prop}
\end{figure}

Figure \ref{increased_prop} shows the results, a reference line at y-intercept of 1 is drawn to aid visual comparisons. Our model performs better on average compared to MICE, irrespective of missing data proportion. Results echo the findings of our main results, where we observe our model performing better than MICE on average of $>85\%$ cases.

\subsection{Impact on final analysis}
Main goal of imputing missing data is to generate complete datasets that can be used for analytics. While imputation accuracy provides us with a measure of how close the imputed dataset is to the complete dataset, we still do not know how well inter-variable correlations are preserved. Which severely impacts end of the line analytics. To check the imputation quality in relation to a dataset's overall structure and to quantify the impact of imputation on end of the line analytics, we use all imputed datasets as the input to classification/regression models based on random forest with 5 times 5 fold cross validation. The task is to use the target variable from all datasets and store the classification accuracy/RMSE for each dataset imputed using our model and MICE. Higher values for classification accuracy and lower RMSE will signify a better preserved predictive dataset structure. We calculate mean accuracy/RMSE from all five runs of multiple imputation. Datasets with data MNAR (uniform and random) are used as MNAR datasets pose greatest challenges for imputation. 

Results in Table \ref{endline} show that multiple imputation using our model provides higher predictive power for end of the line analytics compared to MICE imputed data. The difference even more significant when data are MNAR uniform compared to when data are MNAR random.

\begin{table}[t]
\centering
\caption{Average accuracy and RMSE estimates for end of the line analytics using random forest on imputed datasets. As we have used multiple imputation, results are averaged over all imputed datasets. * signifies where RMSE is reported because target variable is numeric, hence lower values the better. All other datasets report average classification accuracy, higher the better. For dataset VW, as MICE was unable to impute a full dataset, end of the line analytics is not possible.}
\label{endline}
\begin{tabular}{|l|l|l|l|l|l|}
\hline
                       & Data & \multicolumn{2}{l|}{\phantom{}Uniform missingness}            & \multicolumn{2}{l|}{\phantom{}Random missingness}            \\ \hline
                       &      & DAE                & MICE                & DAE                & MICE               \\ \hline
\
\multirow{15}{*}{MNAR} & BH*     & 3.9              & 4.5                   & 3.7                & 4.1                  \\ \cline{2-6}
                       & BC     & 96.0                      & 96.0                  & 97.0               & 96.1                 \\ \cline{2-6}
                       & DN     & 91.6             & 87.6                  & 93.3               & 93.7         \\ \cline{2-6}
                       & GL     & 70.2             & 64.1                  & 74.6               & 70.4                  \\ \cline{2-6}
                       & HV     & 98.5                      & 99.2         & 95.0               & 98.2        \\ \cline{2-6}
                       & IS     & 90.5             & 86.6                  & 90.7               & 90.3        \\ \cline{2-6}
                       & ON*     & 4                         & 3.8          & 3.6                & 4.2         \\ \cline{2-6}
                       & SL     & 90.0             & 80.9                  & 89.6               & 89.6                  \\ \cline{2-6}
                       & SR*     & 6.6              & 8.1                   & 6.9                & 6.4                 \\ \cline{2-6}
                       & ST     & 86.4             & 80.8                  & 76.5               & 72.9         \\ \cline{2-6}
                       & SN     & 90.9             & 70.5                  & 85.3               & 84.2      \\ \cline{2-6}
                       & SB     & 72.6             & 62.7                  & 73.7               & 77.1        \\ \cline{2-6}
                       & VC     & 74.7             & 63.8                  & 72.7               & 70.6        \\ \cline{2-6}
                       & VW     & 93.9             & NA                  & 77.7               & NA       \\ \cline{2-6}
                       & ZO     &   99.9           & 98.5                  & 99.9               & 99.9         \\ \hline
\end{tabular}
\end{table}

\section{Conclusion} \label{sec:conculsion}
We have presented a new method for multiple imputation based on deep denoising autoencoders. We have shown that our proposed method outperforms current state-of-the-art using various real life datasets and missingness mechanisms. We have shown that our model performs well, even with small sample sizes, which is thought to be a hard task for deep architectures. In addition to not requiring a complete dataset for training, we have shown that our proposed model improves end of the line analytics.

\bibliographystyle{plain}
\bibliography{pakdd}
\end{document}